%% file: Paper.tex
\documentclass{esannV2}
\usepackage{graphicx}
\usepackage{subfig}
\usepackage[latin1]{inputenc}
\usepackage{amssymb,amsmath,array,bm}
\usepackage{authblk}
\usepackage{lipsum}
\usepackage{multicol,multirow}
\usepackage{wrapfig}
\newcolumntype{C}[1]{>{\centering\arraybackslash}p{#1}}

\begin{document}
    
    \input{Misc/Info.tex}

    \maketitle
        
    \input{Body/Abstract.tex}

    \input{Body/Introduction.tex}
    \input{Body/Methodology.tex}
    \input{Body/Results.tex}

    \input{Body/Conclusion.tex}
    
    \begin{footnotesize}
        \bibliographystyle{unsrt}
        \bibliography{Misc/theBiblio.bib}    
    \end{footnotesize}
    
\end{document}

%% file: Misc/Info.tex
\title{Short-term forecasting of Amazon rainforest fires based on ensemble decomposition model}

\author{Ramon Gomes da Silva$^{1}$, Matheus Henrique Dal Molin Ribeiro$^{1,2}$, Viviana Cocco Mariani$^{3,4}$ and Leandro dos Santos Coelho$^{1,3}$
%
\thanks{The authors would like to thank National Council of Scientific and Technological Development of Brazil - CNPq (Grants number: 303906/2015-4-PQ and 404659/2016-0-Univ), PRONEX `\textit{Funda\c{c}\~ao Arauc\'aria}' 042/2018, and \textit{Coordena\c{c}\~ao de Aperfei\c{c}oamento de Pessoal de N\'ivel Superior} - Brazil -- CAPES (Finance Code: 001) for financial support of this work.}
%
\vspace{.3cm}\\
%
$^1$ \small{Pontifical Catholic University of Parana, Industrial \& Systems Engineering Graduate Program, Curitiba, PR -- Brazil}
%
\vspace{.1cm}\\
$^2$ \small{Federal Technological University of Parana, Department of Mathematics, Pato Branco, PR -- Brazil}
%
\vspace{.1cm}\\
$^3$ \small{Federal University of Parana, Dept. of Electrical Eng., Curitiba, PR -- Brazil}
%
\vspace{.1cm}\\
$^4$ \small{Pontifical Catholic University of Parana, Mechanical Engineering Graduate Program, Curitiba, PR -- Brazil}
}

%% file: Body/Abstract.tex
\begin{abstract}
Accurate forecasting is important for decision-makers. Recently, the Amazon rainforest is reaching record levels of the number of fires, a situation that concerns both climate and public health problems. Obtaining the desired forecasting accuracy becomes difficult and challenging. In this paper were developed a novel heterogeneous decomposition-ensemble model by using Seasonal and Trend decomposition based on Loess in combination with algorithms for short-term load forecasting multi-month-ahead, to explore temporal patterns of Amazon rainforest fires in Brazil. The results demonstrate the proposed decomposition-ensemble models can provide more accurate forecasting evaluated by performance measures. Diebold-Mariano statistical test showed the proposed models are better than other compared models, but it is statistically equal to one of them.
\end{abstract}

%% file: Body/Introduction.tex
\section{Introduction \label{INT}}

\begin{wrapfigure}{R}{0.5\textwidth}
  \vspace{-20pt}
  \begin{center}
    \includegraphics[width=0.48\textwidth]{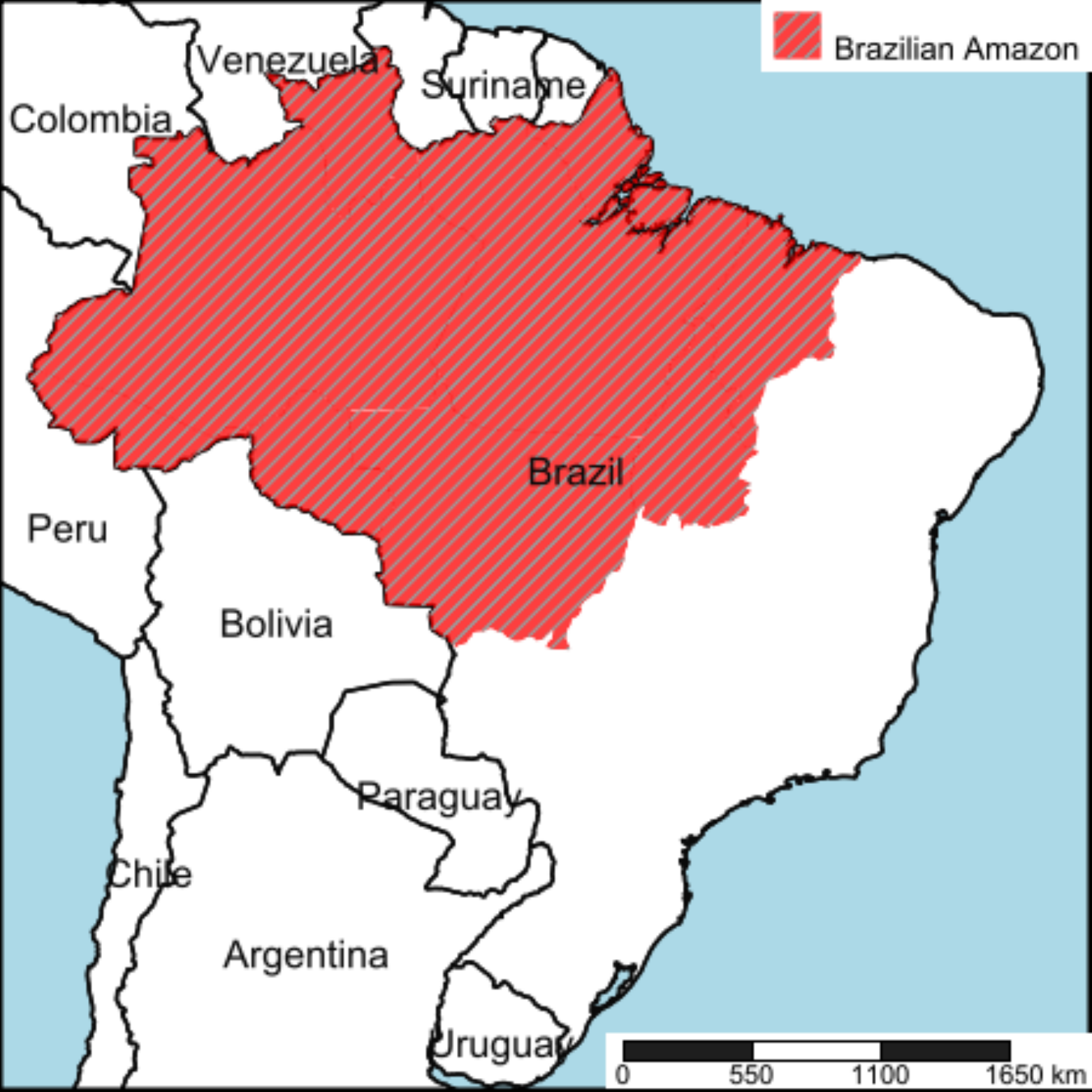}
  \end{center}
  \vspace{-20pt}
  \caption{Brazilian Amazon rainforest area\label{fig:map}}
  \vspace{-10pt}
\end{wrapfigure}

The Amazon rainforest represents over half of the planet's remaining rainforests and comprises the largest and most biodiverse tract of tropical rainforest in the world. Brazil contains the majority part of the Amazon rainforest, with 60\% of it (see illustration in Figure~\ref{fig:map}), followed by Peru with 13\%, Colombia with 10\%, and with minor amounts in Venezuela, Ecuador, Bolivia, Guyana, Suriname, and French Guiana.
In the last years, the fire spots in Amazon has brought attention to world climate discussions and also public health problems in the region \cite{machadosilva2020drought}. Therefore, many studies have been developed discussing the impacts of the fire and how to control it, however, there is a lack of studies to forecast the number of fire spots in the region. Seasonal and Trend decomposition based on Loess (STL) showed to be effective and popular in time series forecasting, mainly to seasonal series. Moreover, ensemble learning approaches increasing the accuracy and efficiency of the models which learns different data patterns, combining potentialities of each base (weak) model making them efficient. 

In this respect, the objective of this paper is to develop a heterogeneous ensemble learning model by using STL decomposition with machine learning models algorithms to train the components generated by decomposition for fire spots cases forecasting in the Brazilian Amazon rainforest multi-month-ahead (one and two-months ahead). The time series is split into three different components and train each component of the STL decomposition with $k$-Nearest Neighbor ($k$-NN), Multivariate Adaptive Regression Splines (MARS), Support Vector Regression (SVR) with kernel Radial, Boosted Generalized Linear Model (GLMBoost), Cubist and Multi-Layer Perceptron (MLP). Then, the predictions of the components are summed giving one thousand combinations by Grid-Search. The combination chosen for one-month-ahead by its performance is composed by SVR with kernel Radial, MARS, and GLMBoost, respectively, and for two-months-ahead is composed by Cubist, $k$-NN and MLP, respectively.  

The main contributions of this study are: (i) to develop a novel heterogeneous decomposition-ensemble learning model by using STL decomposition combined with machine learning models algorithms; (ii) to compare between heterogeneous and homogeneous decomposition and non-decomposed models to evaluate the performance of the proposed model; (iii) to realize comparisons between the predictions of the models of the multi-step-ahead; and (iv) to present a relevant novel to the environmental field, as well to time series forecasting multi-step-ahead using STL decomposition-ensemble model. 

The remainder of this paper is structured as follows. Section~\ref{MET} describes the dataset adopted and presents the methodology applied in this paper. Section~\ref{RES} shows the results and discussions. Finally, Section~\ref{CONC} concludes the paper.

%% file: Body/Methodology.tex
\section{Material \& Methodology \label{MET}}

The dataset analyzed in this paper refers to active fire spots in Brazilian Amazon rainforest area collected from the database of Fire Program of Brazil's National Institute for Space Research (INPE) \cite{INPE2019}. The dataset consists of 255 observations monthly from June 1998 to August 2019. To determine system lags to create inputs, simulations were conducted varying the lag from 1 to 10, which lag equals to 10 (where the training data started since April 1999) presented better results. Moreover, the dataset was split into training and test sets in the proportion of 70\% and 30\%, respectively. In Table~\ref{tab:summary} is presented a summary of the statistical indicators of the dataset, which are the Maximum (Max), Minimum (Min), Mean, Median and Standard Deviation (Std).

\begin{table}[htb!]
\centering
\scriptsize
\begin{tabular}{llcccccc}
\hline
\multirow{2}{*}{Variable} & \multirow{2}{*}{Samples} & \# of & \multicolumn{5}{c}{Statistical indicator} \\ \cline{4-8} 
 &  & Samples & Max & Min & Mean & Median & Std \\ \hline
Fire spots & All set & 245 & 73141 & 70 & 9427 & 3131 & 13249.01 \\
 & Training set & 171 & 73141 & 70 & 10273 & 3175 & 14788.60 \\
 & Test set & 74 & 36569 & 379 & 7473 & 2792 & 8477.56 \\ \hline
\end{tabular}
\caption{Summary of the statistical indicators of the dataset}
\label{tab:summary}
\end{table}

\begin{figure}[htb!]
    \centering
    \includegraphics[width=0.7\linewidth]{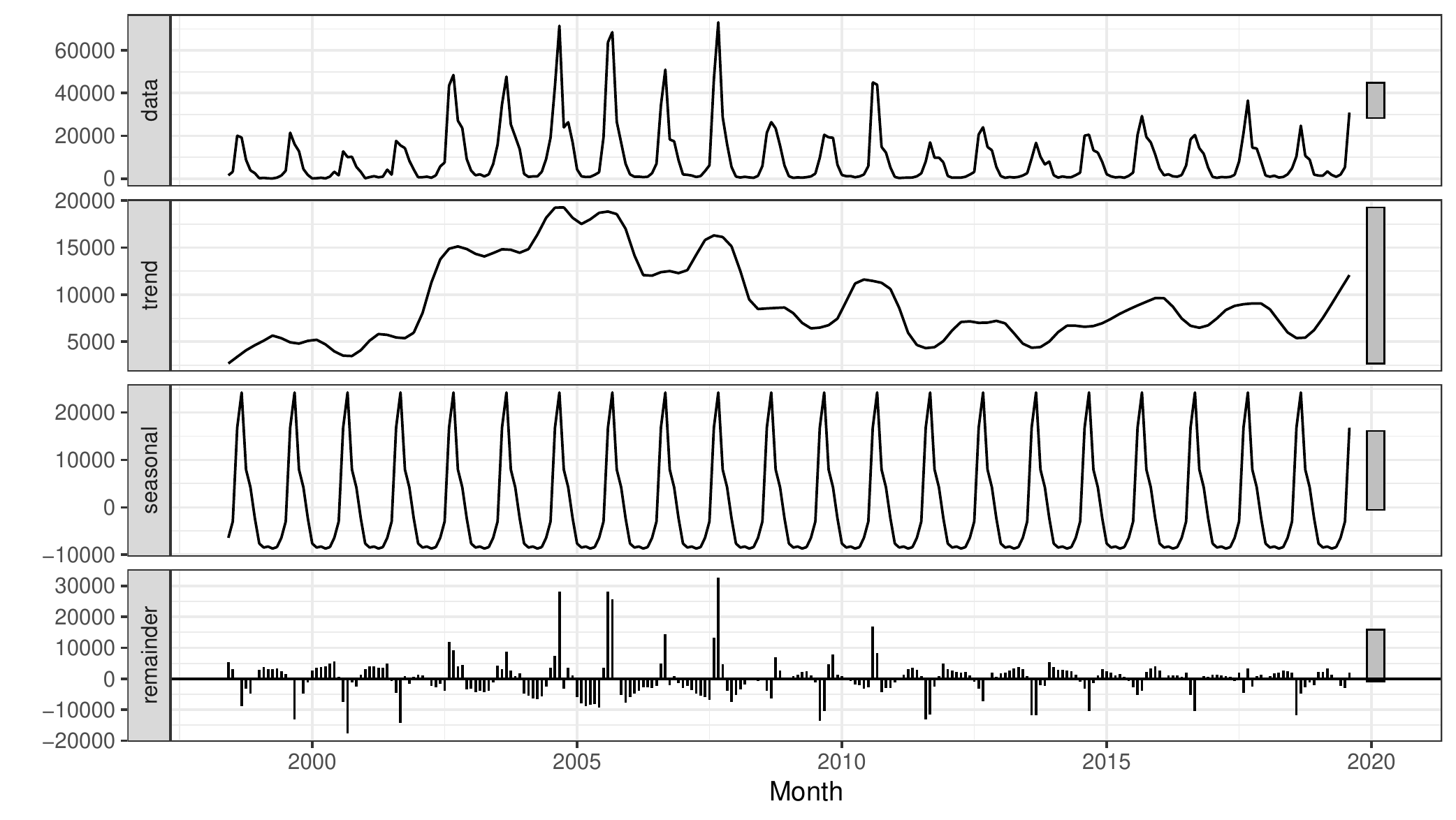}
    \caption{STL decomposition\label{fig:stl}}
\end{figure}

The experiments discussed in this paper were developed using R statistical software. The dataset was shared into three components: seasonal, trend and remainder, based on STL decomposition \cite{cleveland1990stl}, illustrated on Figure~\ref{fig:stl}. 

Each component was trained using the following six models: 
$k$-NN \cite{cover1967nearest},
MARS \cite{friedman1991multivariate},
SVR with kernel Radial \cite{cortes1995support}, 
GLMBoost \cite{buhlmann2003boosting}, 
Cubist \cite{quinlan1992proceedings} and 
MLP \cite{Rumelhart1987}. 
A time-slice validation and Principal Components Analysis (PCA) \cite{jackson1991user} pre-processing were applied in the training process. For the simulations with STL decomposition, PCA needed 6 components to capture 95 percent of the variance, while for simulations without STL decomposition PCA needed 7 components to capture 95 percent of the variance of the signal. 

To forecast one \eqref{eq:eqOSA} and two-months-ahead \eqref{eq:eqTSA} the applied structures are defined as,
\begin{eqnarray}
     \hat{y}(t+h) &=& f\left\{y(t+h-1),y(t+h-2),\ldots,y(t+h-n_y) \right\},                            \label{eq:eqOSA}\\
    \hat{y}(t+h) &=& f\left\{\hat{y}(t+h-1),y(t+h-2),\ldots,y(t+h-n_y)\right\}, \label{eq:eqTSA}
\end{eqnarray}
where $f$ is a function that maps the fire spots data, $\hat{y}(t+h)$ is the forecast fire spots in horizon $h=1,2$ at time $t$ ($1,\ldots, 245$), $y(t+h-1)$, $\hat{y}(t+h-1)$ are the previous observed and predicted fire spots, $y(t+h-n_y)$ is the previous observed fire spots at lag $n_y = 10$.

The predictions obtained from the trained components are recomposed by a simple summation of the extrapolated trend and seasonal components. For one-month-ahead, the chosen combination, due to its performance, was SVR with kernel Radial for the seasonal component, MARS for trend component and GLMBoost for the remainder, denominated STL-Ensemble-1. And for two-months-ahead, the chosen combination was composed by Cubist, $k$-NN and MLP, for seasonal, trend and remainder component, respectively, namely STL-Ensemble-2. 

Moreover, the performance of the proposed STL-Ensemble-1 is compared to the combination with all components trained using only MARS, SVR with kernel radial, and GLMBoost namely STL-MARS, STL-SVR, and STL-GLMBoost, and it is compared with the same algorithms applied in dataset without decomposition. In the same way, STL-Ensemble-2 is compared with the combinations using $k$-NN, Cubist, and MLP, namely STL-KNN, STL-CUBIST, and STL-MLP, as well as the models without decomposition. In order, these comparisons were made to show that the heterogeneous decomposition performed better than homogeneous and even without decomposing.

To evaluate the performance of the various forecasting ensemble models, including the proposed model, measures were adopted. The metrics used in this study are relative root mean square error (RRMSE)~\eqref{eq:RRMSE} and coefficient of determination ($R^2$)~\eqref{eq:R2}, where, $y_i$ is the observed value, $\hat{y}_i$ is the predicted value and $\overline{y}$ is the mean of $y_i$, that are identified by, 
\begin{multicols}{2}
\small
\begin{equation}
    \text{RRMSE}=\frac{\sqrt{\frac{1}{n} \sum_{i=1}^{n}\left(y_i-\hat{y}_i\right)^{2}}}{\frac{1}{n} \sum_{i=1}^{n} y_i},
\label{eq:RRMSE}
\end{equation} 

\begin{equation}
    R^{2}=1-\frac{\sum_{i=1}^{n}\left[y_{i}-\hat{y}_{i}\right]^{2}}{\sum_{i=1}^{n}\left[y_{i}-\overline{y}\right]^{2}}
\label{eq:R2}
\end{equation}
\end{multicols}

Moreover, a Diebold-Mariano test was conducted to evaluate the statistical difference of the proposed models from the other models. 

%% file: Body/Results.tex
\section{Results \label{RES}}
\begin{figure}[htb!]
    \centering
    \includegraphics[width=0.65\linewidth]{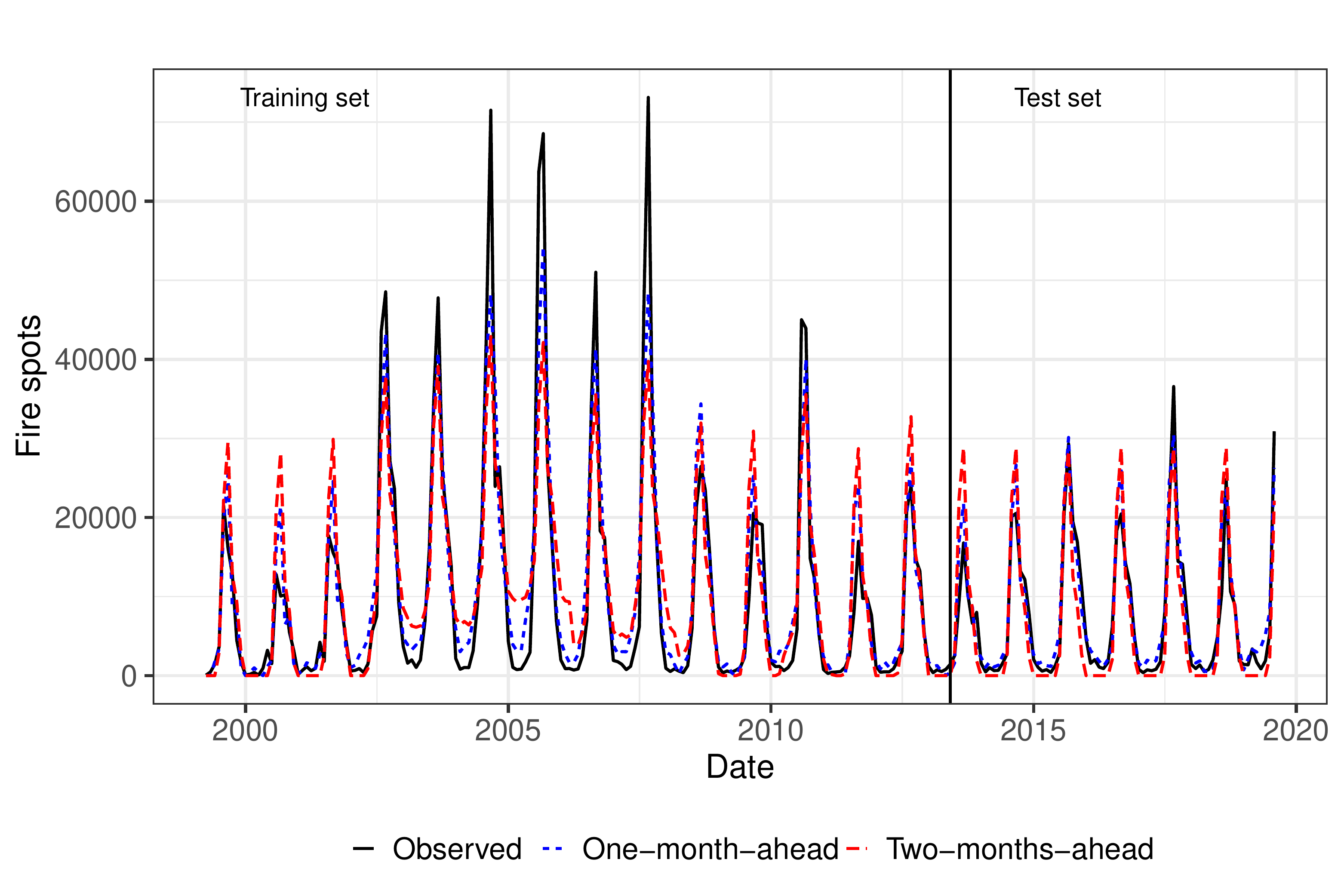}
    \caption{Fire spots prediction for STL-Ensemble model}
    \label{fig:STL-Ensemble}
\end{figure}

The hyperparameters of the models used in this paper, presented in Table~\ref{tab:hyper}, were defined by using a Grid Search. In Figure~\ref{fig:STL-Ensemble} are depicted the STL-Ensemble-1 model fire spots prediction values (blue dotted line), STL-Ensemble-2 model fire spots prediction values (red dotted line) with the observed time series (black line). The analysis reveals that in both training and test datasets the models learned and performed accurately following series trends. Furthermore, as reported in Table~\ref{tab:measures} the performance measures also reflect which in the horizon of one-month-ahead STL-Ensemble-1 model results (in bold) clearly performed better than models cited in Section~\ref{MET}, and in the horizon of two-months-ahead STL-Ensemble-2 (in bold) had a better performance compared to the other models, both exhibiting the importance of combined methodology. Diebold-Mariano test (DM) was conducted comparing the STL-Ensemble-1 with each of the other models of its horizon, the same for STL-Ensemble-2, presented in Table~\ref{tab:DMtest}. DM test shows that STL-Ensemble-1 and STL-Ensemble-2 are better than the models compared, but STL-Ensemble-1 error is statistically equal to STL-MARS error once its $p$-value is greater than 0.1.

\begin{table}[htb!]
\centering
\resizebox{\textwidth}{!}{%
\begin{tabular}{clccc|clcc}
\hline
\multirow{2}{*}{Model} & \multirow{2}{*}{Components} & \multicolumn{3}{c|}{Control Hyperparameters} & \multirow{2}{*}{Model} & \multirow{2}{*}{Components} & \multicolumn{2}{c}{Control Hyperparameters} \\ \cline{3-5} \cline{8-9} 
 &  & \multicolumn{3}{c|}{\# of Neighbors} &  &  & \multicolumn{2}{c}{\# of Boosting Iterations} \\ \hline
$k$-NN & Seasonal & \multicolumn{3}{c|}{9} & GLMBoost & Seasonal & \multicolumn{2}{c}{250} \\
 & Trend & \multicolumn{3}{c|}{5} &  & Trend & \multicolumn{2}{c}{50} \\
 & Remainder & \multicolumn{3}{c|}{11} &  & Remainder & \multicolumn{2}{c}{100} \\
 & Nondecomposed & \multicolumn{3}{c|}{7} &  & Nondecomposed & \multicolumn{2}{c}{250} \\ \hline
 &  & \# of Terms & \multicolumn{2}{c|}{Product Degree} &  &  & \# of Committees & \# of Instances \\ \hline
MARS & Seasonal & 9 & \multicolumn{2}{c|}{1} & Cubist & Seasonal & 1 & 5 \\
 & Trend & 3 & \multicolumn{2}{c|}{1} &  & Trend & 1 & 5 \\
 & Remainder & 9 & \multicolumn{2}{c|}{1} &  & Remainder & 1 & 0 \\
 & Nondecomposed & 2 & \multicolumn{2}{c|}{1} &  & Nondecomposed & 10 & 0 \\ \hline
 &  & Sigma & Cost & Kernel &  &  & \multicolumn{2}{c}{\# of Hidden Units} \\ \hline
SVR & Seasonal & 0.0996 & 4 & Radial & MLP & Seasonal & \multicolumn{2}{c}{9} \\
 & Trend & 0.9212 & 2 & Radial &  & Trend & \multicolumn{2}{c}{9} \\
 & Remainder & 0.0881 & 0.25 & Radial &  & Remainder & \multicolumn{2}{c}{1} \\
 & Nondecomposed & 0.2105 & 2 & Radial &  & Nondecomposed & \multicolumn{2}{c}{1} \\ \hline
\end{tabular}%
}
\caption{Control hyperparameters for the models}
\label{tab:hyper}
\end{table}

\begin{table}[htb!]
\centering
\resizebox{\textwidth}{!}{%
\begin{tabular}{ccccccccc}
\hline
Horizon & Metric & STL-Ensemble-1 & STL-MARS & STL-SVR & STL-GLMBoost & MARS & SVR & GLMBoost \\ \hline
One-month- & RRMSE & \textbf{0.3197} & 0.3903 & 0.3470 & 0.5961 & 0.6661 & 0.6540 & 0.6395 \\
ahead & $R^2$ & \textbf{0.9132} & 0.8957 & 0.9000 & 0.6818 & 0.5911 & 0.6742 & 0.5542 \\ \hline
 &  & STL-Ensemble-2 & STL-KNN & STL-CUBIST & STL-MLP & $k$-NN & CUBIST & MLP \\ \hline
Two-months- & RRMSE & \textbf{0.6311} & 1.4046 & 3.5327 & 1.7753 & 0.6641 & 0.7482 & 0.8575 \\
ahead & $R^2$ & \textbf{0.8186} & 0.7236 & 0.5425 & 0.3504 & 0.6490 & 0.5856 & 0.0581 \\ \hline
\end{tabular}%
}
\caption{Performance measures of the models on test set}
\label{tab:measures}
\end{table}

\begin{table}[htb!]
\centering
\resizebox{\textwidth}{!}{%
\begin{tabular}{cccccccc}
\hline
Horizon &  & \begin{tabular}[c]{@{}c@{}}STL-Ensemble-1\\ vs\\ STL-MARS\end{tabular} & \begin{tabular}[c]{@{}c@{}}STL-Ensemble-1\\ vs\\ STL-SVR\end{tabular} & \begin{tabular}[c]{@{}c@{}}STL-Ensemble-1\\ vs\\ STL-GLMBoost\end{tabular} & \begin{tabular}[c]{@{}c@{}}STL-Ensemble-1\\ vs\\ MARS\end{tabular} & \begin{tabular}[c]{@{}c@{}}STL-Ensemble-1\\ vs\\ SVR\end{tabular} & \begin{tabular}[c]{@{}c@{}}STL-Ensemble-1\\ vs\\ GLMBoost\end{tabular} \\ \hline
One-month- & DM & -0.8191 & -1.8836 & -3.2631 & -3.4462 & -2.5824 & -3.8813 \\
ahead & $p$-value & 0.4154 & 0.0636 & 0.0016 & 0.0009 & 0.0118 & 0.00024 \\ \hline
Horizon &  & \begin{tabular}[c]{@{}c@{}}STL-Ensemble-2\\ vs\\ STL-KNN\end{tabular} & \begin{tabular}[c]{@{}c@{}}STL-Ensemble-2\\ vs\\ STL-CUBIST\end{tabular} & \begin{tabular}[c]{@{}c@{}}STL-Ensemble-2\\ vs\\ STL-MLP\end{tabular} & \begin{tabular}[c]{@{}c@{}}STL-Ensemble-2\\ vs\\ $k$-NN\end{tabular} & \begin{tabular}[c]{@{}c@{}}STL-Ensemble-2\\ vs\\ CUBIST\end{tabular} & \begin{tabular}[c]{@{}c@{}}STL-Ensemble-2\\ vs\\ MLP\end{tabular} \\ \hline
Two-months- & DM & -6.1844 & -9.3799 & -7.8515 & -5.1354 & -8.3402 & -17.1809 \\
ahead & $p$-value & 9.8633e-10 & 6.4036e-20 & 1.2964e-14 & 3.5252e-07 & 3.1566e-16 & 1.1203e-56 \\ \hline
\end{tabular}%
}
\caption{Diebold-Mariano tests}
\label{tab:DMtest}
\end{table}

%% file: Body/Conclusion.tex
\section{Conclusion \label{CONC}}
In this paper, heterogeneous decomposition-ensemble (STL-Ensemble-1 and STL-Ensemble-2) models were proposed to short-term forecasting of Brazilian Amazon rainforest fire cases multi-month-ahead. The STL-Ensemble-1 model was compared with six different approaches, namely as STL-MARS, STL-SVR, STL-GLMBoost, MARS, SVR and GLMBoost, while STL-Ensemble-2 model was compared to others, namely as STL-KNN, STL-CUBIST, STL-MLP, $k$-NN, Cubits, and MLP. The models were evaluated by RRMSE and $R^2$ performance measures, and our proposed models yield better forecast accuracy than other models. Further, the Diebold-Mariano test was conducted and indicated that STL-Ensemble-1 and STL-Ensemble-2 were more accurate than the compared, yet the STL-Ensemble-1 error is statistically equal to STL-MARS error. Finally, we can conclude that our proposed models can help government agency policies to predict and prevent fire spots in Amazon rainforest. As future research propositions are intended to perform a multi-step-ahead forecast with a wider horizon, using different algorithms and decomposition methods.